\documentclass[11pt,a4paper]{article}
\usepackage{authblk}
\usepackage[hyperref]{acl2021}
\usepackage{times}
\usepackage{latexsym}

\usepackage{amsmath}
\usepackage{microtype}
\usepackage{graphicx}
\usepackage{multirow}
\usepackage{tabu}
\usepackage{longtable}
\usepackage{amssymb}
\usepackage{booktabs}
\usepackage{enumitem}
\usepackage{textcomp}
\graphicspath{ {./pictures/} }

\usepackage{parskip}
\usepackage{xr}
\aclfinalcopy 
\def\aclpaperid{56} %  Enter the acl Paper ID here

\title{A Single Example Can Improve Zero-Shot Data Generation}
\author[1]{\bf{Pavel Burnyshev}} 
\author[1,2]{\bf{Valentin Malykh}}
\author[1]{\bf{Andrey Bout}}
\author[1,3]{\bf{Ekaterina Artemova}}
\author[1]{\bf{Irina Piontkovskaya}}
\affil[1]{Huawei Noah's Ark Lab, Moscow, Russia}
\affil[2]{Kazan Federal University, Kazan, Russia}
\affil[3]{HSE University, Moscow, Russia}
\affil[ ]{\textit {\{burnyshev.pavel, malykh.valentin, bout.andrey, artemova.ekaterina, piontkovskaya.irina\}@huawei.com}}
\begin{document}
\maketitle
\begin{abstract}

Sub-tasks of intent classification, such as robustness to distribution shift, adaptation to specific user groups and personalization, out-of-domain detection, require extensive and flexible datasets for experiments and evaluation. As collecting such datasets is time- and labor-consuming, we propose to use text generation methods to gather datasets. The generator should be trained to generate utterances that belong to the given intent. 
We explore two approaches to generating task-oriented utterances. In the {\bf zero-shot approach}, the model is trained to generate utterances from seen intents and is further used to generate utterances for intents unseen during training. In the {\bf one-shot approach}, the model is presented with a single utterance from a test intent.  We perform a thorough automatic, and human evaluation of the dataset generated utilizing two proposed approaches. Our results reveal that the attributes of the generated data are close to original test sets, collected via crowd-sourcing.
% We will publish the generated dataset and the code in open access, which allows anyone to generate the new data of desirable properties.

\end{abstract}

\section{Introduction} 

Training dialogue systems used by virtual assistants in task-oriented applications requires large annotated datasets. The core machine learning task to every dialogue system is {\it intent detection}, which aims to detect what the intention of the user is. New intents emerge when new applications, supported by the dialogue systems, are launched. However, an extension to new intents may require annotating additional data, which may be time-consuming and costly. What is more, when developing a new dialogue system, one may face the cold start problem if little training data is available. Open sources provide general domain annotated datasets, primarily collected via crowd-sourcing or released from commercial systems, such as Snips NLU benchmark \cite{coucke2018snips}. However, it is usually problematic to gather more specific data from any source, including user logs,  protected by the privacy policy in real-life settings. 

For all these reasons, we suggest a learnable approach to create training data for intent detection. We simulate a real-life situation in which no annotated data but rather only a short description of a new intent is available. To this end, we propose to use methods for zero-shot conditional text generation to generate plausible utterances from intent descriptions. The generated utterances should be in line with the intent's meaning.  

% To justify our approach, we utilize both computational and human evaluation. First, we use several fluency, diversity, and semantic consistency metrics to evaluate the quality of the generated data. Second, we conduct a thorough evaluation of the generated data utilizing a crowd-sourcing study. Our results show that the zero-shot generation has its limitation. Sometimes, the meaning of the generated utterances differs from expected, perhaps due to the generation model's lack of world knowledge. To overcome this issue, we propose a novel one-shot adjustment method, which relies at reinforcement learning techniques. Besides, pre-trained language models, that excel at generating human-like utterances and natural language understanding, are core to our approach.  We utilize the GPT-2 model \cite{radford2019language} along with different decoding strategies to generate utterances and the BERT-based reward \cite{zhang2019bertscore} in the one-shot setting.

Our contributions are:
\begin{enumerate}[topsep=0pt,itemsep=-1ex,partopsep=1ex,parsep=1ex]
    \item We propose a zero-shot generation method  to generate a task-oriented utterance from an intent description;
    \item We evaluate the generated utterances and compare them to the original crowd-sourced datasets. The proposed zero-shot method achieves high scores in fluency and diversity as per our human evaluation; 
    \item We provide experimental evidence of a semantic shift  when generating utterances for unseen classes using the zero-shot approach;
    \item We apply reinforcement learning for the one-shot generation to eliminate the semantic shift problem. The one-shot approach retains semantic accuracy without sacrificing fluency and diversity.
\end{enumerate}

\section{Related work}

% This project is based upon the methods for conditional text generation and zero-shot learning. These methods allow us to generate training data for intent classification. 

\paragraph{Conditional language modelling} generalizes the task of language modelling. Given some conditioning context $z$, it assigns probabilities to a sequence of tokens \cite{mikolov2012context}. Machine translation \cite{sutskever2014sequence,cho2014learning} and image captioning \cite{you2016image} are seen as typical conditional language modelling tasks. More sophisticated tasks include text abstractive summarization \cite{nallapati2017summarunner,narayan2019article} and simplification \cite{zhang2017sentence},  generating textual comments to source code \cite{richardson2017code2text} and dialogue modelling \cite{lowe2017training}. Structured data may act as a conditioning context as well. Knowledge base (KB) entries \cite{vougiouklis2018neural} or DBPedia triples \cite{colin2016webnlg} serve as condition to generated plausible factual sentences. Neural models for conditional language modelling rely on encoder-decoder architectures and can be learned both jointly from scratch  \cite{vaswani2017attention} or by fine-tuning pre-trained encoder and decoder models \cite{budzianowski2019hello,lewis2019bart}.

\paragraph{Zero-shot learning (ZSL)} has formed as a recognized training paradigm with neural models becoming more potent in the majority of downstream tasks. In the NLP domain, the ZSL scenario aims at assigning a label to a piece of text based on the label description. The learned classifier becomes able to assign class labels, which were unseen during the training time. The classification task is then reformulated in the form of question answering \cite{levy2017zero} or textual entailment \cite{yin2019benchmarking}. 
%Other approaches leverage linguistic information extracted from the description to train classifiers \cite{srivastava2018zero}.  
Other techniques for ZSL leverage metric learning and make use of capsule networks \cite{du2019investigating} and prototyping networks \cite{yu2019episodebased}. 

\paragraph{Zero-shot conditional text generation} implies that the model is trained in such a way that it can generalize to an unseen condition, for which only a description is provided. A few recent works in this direction show-case dialog generation from unseen domains \cite{zhao2018zero} and question generation from KB's from unseen predicates and entity types \cite{elsahar2018zero}. CTRL \cite{keskar2019ctrl}, pre-trained on so-called control codes, which can be combined to govern style, content, and surface form, provides for zero-shot generation for unseen codes combinations. PPLM  \cite{dathathri2019plug} uses signals, representing the class, e.g., bag-of-words, during inference, and can generate examples with given semantic attributes without pre-training.

\paragraph{Training data generation} can be treated as form of data augmentation, a research direction being increasingly in demand. It enlarges datasets for training neural models and help avoid labor-intensive and costly manual annotation. Common techniques for textual data augmentation include back-translation \cite{sennrich2016improving}, sampling from latent distributions \cite{xia2021pseudo}, simple heuristics, such as synonym replacement \cite{wei2019eda} and oversampling \cite{chawla2002smote}. Few-shot text generation has been applied to natural language generation from structured data, such as tables \cite{chen2020few} and to intent detection data augmentation \cite{xia2021pseudo}. However, these methods are incompatible with ZSL, requiring at least a few labeled examples for the class being augmented. 
An alternative approach suggests to use a  model to generate data for the target class based on task-specific world knowledge \cite{chen2017automatically} and linguistic features \cite{iyyer2018adversarial}.

\paragraph{Deep reinforcement learning (RL)} methods prove to be effective in a variety of NLP tasks. Early works approach the tasks of machine translation \cite{grissom2014don}, image captioning \cite{rennie2017self} and abstractive summarization \cite{paulus2017deep}, assessed with not differentiable metrics.  \cite{wu2020textgail} tries to improve the quality of transformer-derived pre-trained models for generation by leveraging proximal policy optimization. Other applications of deep RL include dialogue modeling
\cite{li2016deep} and open-domain question answering \cite{wang2018r}.

\section{Methods}

Our main goal is to generate plausible and coherent utterances, which relate to unseen intents, leveraging the description of the intent only. These utterances should clearly express the desired intent. For example, if conditioned on the intent {\it ``delivery from the grocery store''} the model should generate an utterance close to  {\it ``Hi! Please bring me milk and eggs from the nearest convenience store''} or similar.

Two scenarios can be used to achieve this goal. In the {\bf zero-shot scenario}, we train the model on a set of seen intents $\mathcal{S}$ to generate utterances. If the generation model generalizes well, the utterances generated for unseen intents  $\mathcal{U}$ are diverse and fluent and retain intents' semantics.  In the {\bf one-shot scenario}, we utilize one utterance per unseen intent $\mathcal{U}$ to train the generation model and learn the semantics of this particular intent.

\subsection{Zero-shot generation}

\begin{figure}[!htp]
\includegraphics[width=\linewidth]{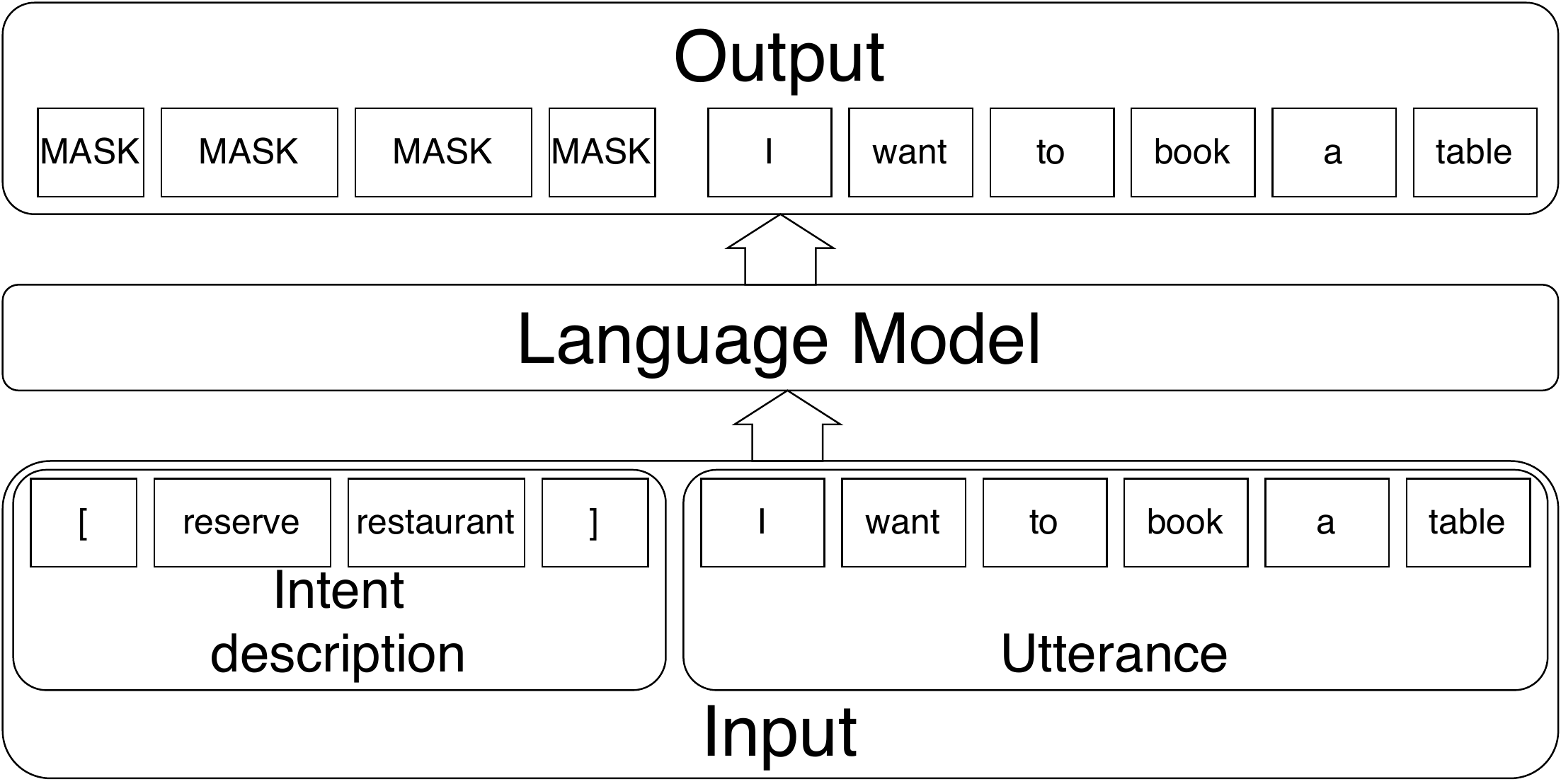}
\caption{Training setup. The input an intent description and an utterance concatenated, the output is the utterance.}
\label{fig:gpttuning}
\end{figure}

Our model as depicted in Figure~\ref{fig:gpttuning}) aims to generate plausible utterances conditioned on the intent description.  We fine-tune the GPT-2 medium model \cite{radford2019language} on task-oriented utterances, collected from several NLU benchmarks (see Section~ \ref{section:data} for more details on the dataset).

Our approach to fine-tuning the GPT-2 model follows \cite{budzianowski2019hello}.  Two pieces of information, the intent description and the utterance are concatenated to form the input. More precisely, the input has the following format: {\it [intent description]}  {\it utterance}. During the training phase, the model is presented with the output obtained from the input by masking the intent description. The output has the following format: \texttt{<MASK>}, $\ldots$, \texttt{<MASK>}  {\it utterance}. The full list of intents is provided in Table 4 in Appendix.

Such input allows the model to pay attention to intent tokens while generating. The standard language modeling objective, negative log-likelihood loss, is used to train the model:
\begin{align*}
\mathcal{L}\left(\theta\right)=-\sum_{i}\sum_{t=1}^{\left|\mathbf{x}^{(i)}\right|} \log p_{\theta}\left(x_{t}^{(i)} |\text{intent}, x_{<t}^{(i)}\right).
\end{align*}

We fine-tuned the model for one epoch to avoid over-fitting. Otherwise, the model tends to repeat redundant semantic constructions of the input utterances.  At the same time, a bias towards the words from the training set gets formed. The parameters of the training used were set to the following values: batch size equals to $32$,  learning rate equals to $5$e-$5$, the optimizer chosen is Adam \cite{kingma2015adam} with default parameters.

\subsection{One-shot Generation}
{\bf Motivation.} The zero-shot approach to conditional generation may degrade or even fail if (i) the intent description is too short to properly reflect the semantics of the intent, (ii) the intent description is ambiguous or contains ambiguous words.  Produced utterances may distort the initial meaning of the intent or be meaningless at all. The model may generate an utterance {\it ``Count the number of people in the United States''} for the intent  ``calculator'', or  {\it ``Add a book by Shakespeare to the calendar''} for a ``book reading'' service. Although such examples can be treated not as outliers but rather as real-life whimsical utterances, this is not the desired behavior for the generation model. We address this phenomenon as \emph{Semantic Shift} and provide experimental evidence of it in Section~\ref{sec:SemanticShift}.

Based on these observations, we hypothesize that the problem could be solved if we provide a single training example to improve models' generalization abilities. A single example can give the model a clue about what the virtual assistant can do with books and which entities our calculator is designed to calculate by gaining better world knowledge. For this purpose, we are moving from the zero-shot to the one-shot setting. We propose a method for improving zero-shot generation by leveraging just one example.

Our approach is inspired by the recent  TextGAIL \cite{wu2020textgail} approach. It addresses the problem of exposure bias in pre-trained language models and proposes a GAN-like style scheme for fine-tuning GPT-2 to produce appropriate story endings using a reinforcement algorithm. As a reward, TextGAIL uses a discriminator output trained to distinguish real samples from generated samples. 
As we are limited in using learnable discriminators because of the lack of training data, we propose an objective function based on a similarity score. Our objective function produces utterances, which are close to the reference example.  At the same time, it forces the model to generate more diverse and plausible utterances. Table 5 in Appendix provides reference examples used for the one-shot generation method.

\noindent{\bf Method.} After zero-shot fine-tuning, we perform a one-shot model update for each intent separately. We perform several steps of the Proximal Policy Optimization algorithm \cite{schulman2017proximal} with the objective function described further.

\noindent{\bf Reward.} Our reward function is based on BERTScore \cite{zhang2019bertscore}, which serves as the measure of contextual similarity between generated sentences and the reference example. BERTScore correlates better with human judgments than other existing metrics, used to control semantics of generated texts and detect paraphrases. Given a reference and a candidate sentence, we embed them using RoBERTa model \cite{liu2019roberta}. The BERTScore F1 calculated on top of these embeddings is used as a part of the final reward.

It is not enough to reward the model only for the similarity of the generated utterance to the reference one. If so, the model tends to repeat the reference example and receives the maximal reword. We add the negative sum of frequencies of all $n$-grams in the utterance to the reward function, forcing the model to generate less frequent sequences. 

Given an intent $I$ and a reference example $x_{\text{ref}}^I$, the reward for the sentence $x$ is calculated by the formula:
\begin{align*}
R_I(x) &= R_{sim}(x_{\text{ref}}^I, x) + R_{div}(x) && \\
R_{sim}(x_{\text{ref}}^I, x) &= \text{BERTScore}(x_{\text{ref}}^I, x)&& \\
R_{div}(x) &= \sum\limits_{s \in \text{n-grams}(x)}(-\nu_s)&& 
\end{align*}
where $\nu_s$ is the $n$-gram frequency, calculated from all the generated utterances inside one batch. 

\noindent{\bf Objective function.} First, we plug this reward into standard PPO objective function, getting intent-specific term  $L^{\text {policy }}_{I}(\theta)$. Following the TextGAIL approach, we add $\mathbf{KL}$ divergence with the model without zero-shot fine-tuning to prevent forgetting the information from the pre-trained model. We add an entropy regularizer, making the distribution smoother, which leads to more diverse and fluent sentences. According to our experiments, this term helps avoid similar prefixes for all generated sentences as $n$-gram reward only does not cope with this issue. The final generator objective for maximization in the one-shot scenario for the intent $I$ can be written as follows:

% \begin{equation}
$
\begin{aligned}
%L(I;\theta) &= \mathbb{E}_{x \sim p_{\theta}(\cdot|I)}[L^{\text{policy}}_I(\theta) - \alpha \mathbf{KL}(p_{\theta}\mid q) + \\ &+ \beta \mathbf{H}(p_{\theta})],
%L(I;\theta) &= \mathbb{E}_{x \sim p_{\theta;I}}[L^{\text{policy}}_I(\theta) - \alpha \mathbf{KL}(p_{\theta;I}\mid q) + \\ &+ \beta \mathbf{H}(p_{\theta;I})],
%EL: можно так, это будет правильный лосс для одного примера
L(I;\theta) =& L^{\text{policy}}_I(\theta)+\hat{\mathbb{E}}_t\large{[} \beta\mathbf{H}(p_{\theta;I}(\cdot | s_t))\large\\ -& \alpha\mathbf{KL}[p_{\theta;I}(\cdot | s_t), q(\cdot | s_t)]{]},
\end{aligned}
% \end{equation}
$
%\hat{\mathbb{E}}_t[\dots] -  indicates the empirical average over an utterance's tokens

\noindent where $s_t$ is intent description, $p_{\theta;I}$ is the conditional distribution $p_{\theta}(\cdot|I)$(distribution, derived from model with updates from PPO policy), $q$ is an unconditional LM distribution, calculated by GPT-2 language model without fine-tuning. The entropy and $\mathbf{KL}$ are calculated per each token, while the $L^{\text{policy}}$ term is calculated for the whole sentence.

\subsection{Decoding strategies}
Recent studies show that a properly chosen decoding strategy significantly improves consistency and diversity metrics and human scores of generated samples for multiple generation tasks, such as story generation \cite{holtzman2019curious}, open-domain dialogues, and image captioning \cite{Ippolito2019comparison}. However, to the best of our knowledge, no method proved to be a one-size-fits-all one. We perform experiments with several decoding strategies, which improve diversity while preserving the desired meaning. We perform an experimental evaluation of different decoding parameters.

\noindent {\bf Beam Search}, a standard decoding mechanism,
keeps the top $b$ partial hypotheses
at every time step and eventually chooses the hypothesis that has the overall highest probability.

\noindent {\bf Random Sampling (top-$k$)} \cite{fan2018hierarchical} greedily samples at each time step   one of the top-$k$ most likely tokens in the distribution.

\noindent {\bf Nucleus Sampling (top-$p$)} \cite{holtzman2019curious} samples from the most likely tokens whose cumulative probability does not exceed $p$. 

\noindent {\bf Post Decoding Clustering} \cite{Ippolito2019comparison} (i) clusters generated samples using BERT-based similarity and (ii) selects samples with the highest probability from each cluster. It can be combined with any decoding strategy.

\section{Performance evaluation}

We use several quality metrics to assess the generated data: (i) we use multiple fluency and diversity metrics, (ii) we account for the performance of the classifiers trained on the generated data. 

\noindent{\bf Fluency.} We consider fluency dependent upon the number of spelling and grammar mistakes: the utterance is treated as a fluent one if there are no misspellings and no grammar mistakes. We utilize LanguageTool \cite{milkowski2010developing}, a free and open-source grammar checker, to check spelling and correct grammar mistakes.

\noindent{\bf Diversity.} Following \cite{Ippolito2019comparison}, we consider two types of diversity metrics:

$Dist\mbox{-}k$ \cite{li2016diversity} is the total number of distinct $k$-grams divided by the total number of produced tokens in all of the utterances for an intent;

$Ent\mbox{-}k$ \cite{zhang2018generating} is an entropy of $k$-grams distribution. This metric takes into consideration that infrequent $k$-grams contribute more to
diversity than frequent ones.

\noindent{\bf Accuracy.}  After we obtain a large amount of generated data, we train a RoBERTa-based classifier \cite{liu2019roberta} to distinguish between different intents, based on the generated utterances. As usual, we split the generated data into two parts so that the first part is used for training, and the second part serves as the held-out validation set to compute the classification accuracy $acc_{clsf}$. High  $acc_{clsf}$ values mean that the intents are well
distinguishable, and the utterances that belong to the same intent are semantically consistent.

\noindent{\bf Human evaluation} We perform two crowd-sourcing studies to evaluate the quality of generated utterances, which aim at the evaluation of semantic correctness and fluency. 

First, we asked crowd workers to evaluate semantic correctness.  We gave crowd workers an utterance and asked them to assign one of the four provided intent descriptions; a correct option was among them (i.e., the one used to generate this very utterance). For the sake of completeness, we added a fifth option, ``none of above''. We assess the results of this study by two metrics, accuracy and $recall@4$. Accuracy $acc_{crowd}$ measures the number of correct answers, while $recall@4$ measures the number of answers which are different from the last  ``none of above'' option.

Second, we asked crowd workers to evaluate the fluency of generated utterances. Crowd workers were provided with an utterance and were asked to score it on a Likert-type scale from 1 to 5, where (5) means that the utterance sounds natural, (3) means that the utterance contains some errors, (1) means that it is hard or even impossible to understand the utterance. We assess the results of this study by computing the average score.

\section{Zero-shot generation experiments}
\subsection{Data preparation}
\label{section:data}

{\bf Data for fine-tuning.} We combined two NLU datasets, namely The Schema-Guided Dialogue Dataset (SGD) \cite{rastogi2019towards} and Natural Language Understanding Benchmark (NLU-bench) \cite{coucke2018snips} for the fine-tuning stage.  
Both datasets have a two-level hierarchical structure: they are organized according to services (in SGD) or scenarios (in NLU-Bench). Each service/scenario contains several intents, typically 2-5 intents per high-level class. For example, the service \emph{Buses\_1} is divided into two intents \emph{FindBus} and \emph{BuyBusTickets}.

SGD dataset consists of multi-turn task-oriented dialogues between user and system; each user utterance is labeled by service and intent. We adopted only those utterances from each dialog in which a new intent arose, which means the user clearly announced a new intention. This is a common technique to remove sentences that do not express any intents. As a result, we got three utterances per dialog on average. 

As NLU-Bench consists of user utterances, each marked up with a scenario and intent label, we used it without filtering. Summary statistics of the dataset used is provided in Table~\ref{tab:finetune_data}.

\begin{table}[h]
\centering
  \begin{tabular}{p{2.59cm}p{1.2cm}p{1.2cm}p{1.2cm}}
    \toprule
           & SGD & NLU-bench & Total \\ \midrule
    No. of utterances & 49986 & 25607& 75593  \\
    No. of services & 32& 18& 50  \\
    No. of intents  & 67 & 68& 135  \\
    Total tokens & $\sim$550k & $\sim$170k &   $\sim$720k \\ 
    Unique tokens & $\sim$10.8k & $\sim$8.3k &   $\sim$17.4k \\ \bottomrule
  \end{tabular}
 \caption{The total number of utterances, intents, services and words across datasets and final statistics of our fine-tuning data.}
 \label{tab:finetune_data}
 \end{table}

\begin{table*}[!htp]
\centering
\begin{tabu} to \textwidth {  l | c c c | c c c  }
\toprule
\multicolumn{7}{c}{Zero-shot generation} \\\toprule
\multirow{2}{*}{Decoding strategy} & \multicolumn{3}{ c |}{Automated metrics} &  \multicolumn{3}{ c }{Human evaluation} \\
\cline{2-7}
& $acc_{clsf}$ & $Dist\mbox{-}4$ & $Ent\mbox{-}4$ & $acc_{crowd}$ & $recall@4$ & Fluency score \\
\midrule
Random Sampling ($b=4$) & 0.82 & \bf{0.50} &	\bf{6.20} & 0.63 & 0.87 & 4.77 \\
Nucleus Sampling ($p=0.6$) + PDC &	0.82 &	0.40 &	5.77 &	0.68 &	0.85 &	\bf{4.95} \\ 
Beam Search ($b=3$) + PDC &	0.85 &	0.22 &	4.92 &	0.67 &	0.85 &	4.88 \\
Beam Search ($b=3$) &	0.88 &	0.15 &	4.76 &	0.60 &	0.80 &	4.76 \\
Nucleus Sampling ($p=0.4$) &	0.89 &	0.25 &	4.95 &	0.72 &	0.90 &	4.81 \\
\bottomrule
\multicolumn{7}{c}{One-shot generation} \\\toprule
Nucleus Sampling ($p=0.4$)  &	\bf{0.94} & 	0.39 & 	5.88 & 	\bf{0.78} & 	\bf{0.91} & 	4.86 \\\bottomrule 

% \multicolumn{7}{|c|}{Ensemble} \\\hline
% ? & ? & ? &	? & ? & ? & ? \\\hline

\end{tabu}
\caption{Decoding strategies for zero-shot and one-shot generation. PDC stands for Post Decoding Clustering.}
\label{tab:genComparison}
\end{table*}

\begin{table*}
\centering
\begin{tabu}
to \textwidth {  l | c c c c }
\toprule
 & $acc_{crowd}$ & $recall@4$ & $Dist\mbox{-}4$ & $Ent\mbox{-}4$ \\\midrule
SGD+NLU-bench & 0.83 & 0.95 &  0.53 &	5.92 \\\bottomrule

\end{tabu}
\caption{Evaluation of the test dataset, created by merging and re-splitting two datasets under consideration.}
\label{tab:trueData}
\end{table*}

\noindent{\bf Intent set for generation.} For the evaluation of our generation methods, we created a set of 38 services and 105 intents\footnote{The full list of services and intents in both sets presented in the Appendix} covering the most common requirements of a typical user of a modern dialogue system. The set includes services dedicated to browsing the Internet, adjusting mobile device settings, searching for vehicles, and others. 
To adopt a zero-shot setup, we split the data into train and test sets in the following way. Some of the services are unseen ($s \in \mathcal{U}$), i.e., are present in the test set only. There are no seen services in the train set related to unseen services. The rest of the services are seen, i.e., present in both train and test set ($s \in \mathcal{S}$), but different intents put in train and test sets. For example, \emph{Flight} services are present in the train data and \emph{Plane}  service is used in the test set; from \emph{Music} services, intents \emph{Lookup song} and \emph{Play song} were used for training, and \emph{Create playlist} and \emph{Turn on music} for a testing. To form the intent description for fine-tuning and generation, we join service and intent labels. 

\subsection{Evaluation}
We generated 100 examples per intent using different decoding strategies and their parameters. For the more detailed evaluation, we picked up the generation methods of different decoding strategies that achieved good scores ($acc_{clsf} >80\%$ and $Ent\text{-}4>4$).  For these utterances, we performed a human evaluation of semantic correctness and diversity; Table~\ref{tab:genComparison} compares the decoding strategies according to various quality metrics. For a more detailed evaluation of decoding strategies, see Table 2 in Appendix.

To compare the diversity of human-generated utterances to our generated utterances, we evaluate the fine-tuning dataset with $Ent\mbox{-}4$ and $Dist\mbox{-}4$ metrics. The semantics of generated data is assessed by $acc_{crowd}$ and $recall@4$. We present metrics for this dataset in Table~\ref{tab:trueData}.

\begin{table*}[!htp]
\centering
\begin{tabular}{p{4.8cm}p{4.8cm}p{5.4cm}}
%{p{\linewidth}p{\linewidth}p{\linewidth}}
\toprule
Beam Search (3) & Random Sampling (3) &
Nucleus Sampling (0.98)\\
$Ent\mbox{-}4 = 4.26$ &$Ent\mbox{-}4 = 5.93$ & $Ent\mbox{-}4 = 6.86$\\

\midrule
i need to know what's going on with my phone &i want to see my messages in the phone book & show me a message from jean lee for my favorite apple company \\
i want you show me the message from my phone & show me my most recent messages from my phone number & how can you tell me mike with the message \\
i want you show me my messages on my phone & show me the messages from the device i was using & could you check to see if my friends are in a group that is gossiping \\
i want you to show my messages on my smart phone & show me the message from my friend jane that i sent to her & list all messages in my bbq menu from ausy \\
i want to read a new message from my friend & can you please show me the messages from my phone & just turn on the smart mute this monday night \\ \bottomrule
\end{tabular}
\caption{Utterances, generated by different decoding strategies and the diversity scores of the decoding strategies.}
\label{tab:diversityExamples}
\end{table*}

\subsection{Analysis and model comparison}
\paragraph{Fluency.} 
 
Spell checking results reveal the following issues of the generated utterances. The major issues are related to casing: an utterance may start in lower case, the first-person singular personal pronoun ``I'' is frequently generated in lower case, too. Punctuation issues include missing quotes, question marks, periods, or repeated punctuation marks. Common mistakes are omitting of a hyphen in the word {\it ``Wi-Fi''} and {\it ``e-mail''} and confusing definite and indefinite articles, as well as confusing {\it ``a''/``an''}. These issues are more or less natural to humans and thus do not prevent further use of generated utterances. The only unnatural issues found by LanguageTool are phrase repetition in small numbers ($4$ errors of this type per $10000$ utterances). For examples of fluency issues in generated data, see Table 1 in Appendix. 
% which can be easily fixed by means of  LanguageTool, regular expressions or any other tool.  

\paragraph{Diversity.} Table~\ref{tab:diversityExamples} shows examples of the phrases generated by means of different decoding strategies, conditioning on the intent \emph{Show message}, along with diversity metrics, $Dist$ and $Ent$. Higher $Ent$ and $Dist$ scores indeed correspond to a more diverse decoding strategy. At the same time, extremely high diversity may generate utterances unrelated to the intent, expressing non-clear meaning and lack of common sense.

\begin{figure}[!htp]
\includegraphics[width=\linewidth]{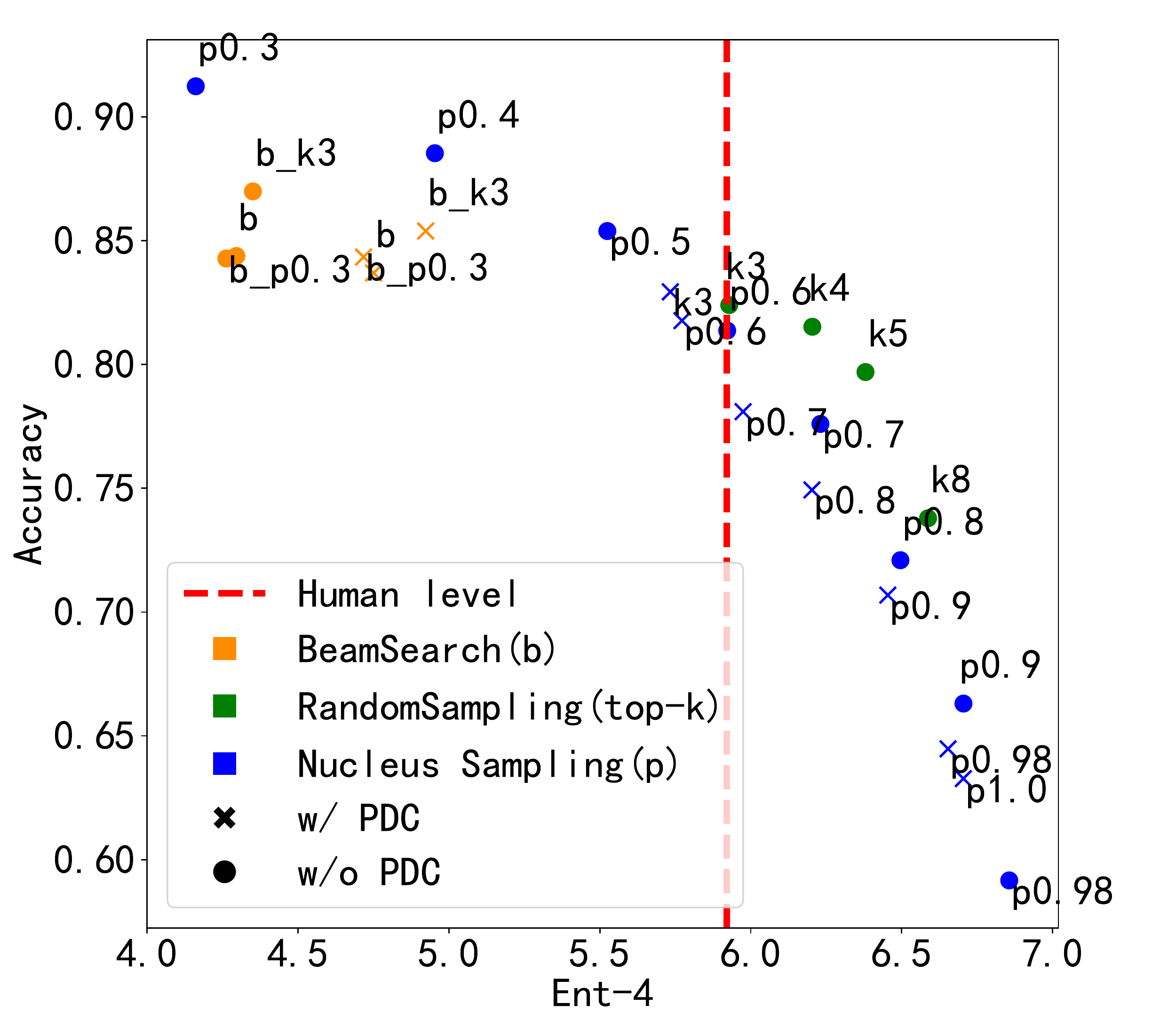}
\caption{The trade-off between diversity ($Ent\mbox{-}4$) and accuracy.}
\label{fig:accuracyDiversity}
\end{figure}

\paragraph{Diversity / Accuracy trade-off.} Figure~\ref{fig:accuracyDiversity} shows the trade-off between the diversity ($Ent\mbox{-}4$) and the accuracy ($acc_{clsf}$) of the generated data.

Every point corresponds to sentences generated using different zero-shot strategies. The human level stands for the diversity and accuracy metrics computed for the test set as is.  The beam search scores are mainly in the top-left corner of the plane, leading to high accuracy and low diversity values. Top-$k$ Random Sampling strategy does not achieve the highest levels of accuracy. Nucleus Sampling can generate datasets with a large range of diversity and accuracy scores, depending on the chosen parameter.  Post-decoding clustering increases diversity for low-diverse decoding strategies and decreases it for high-diverse ones, moving the generator closer to the human level. 
% можно добавить что-то про pdc. Human-level - это на самом деле тотровень, который наблюдался в данных для файн-тюнинга, то есть видно, что pdc корректирует генерацию в направлении training diversity. Это отмечено в оригинальной статье про pdc? 

\paragraph{Two ways to assess accuracy.}  Table~\ref{tab:genComparison} shows that there is no clear correspondence between automated accuracy $acc_{clsf}$ and human accuracy  $acc_{crowd}$. Therefore $acc_{clsf}$ cannot serve as the final measure for the semantic consistency of the generator.
The \emph{Semantic shift} problem cannot be captured by the automated accuracy $acc_{clsf}$: the model generates examples which are consistent inside each class, and classes are well-separated, but the generated examples do not correspond well to the intent descriptions. 

%How to prove that there is the main issue?

%Looks like some problems with low-diverse generators are because they generate data too close to training, and loose semantics, like \cite{That's great, turn it on}, or  \cite{I want to go to San Francisco} for any travel-related request. Maybe it could be checked by BERTScore with fine-tune data?

%We see that human data accuracy is not very high: $0.69$. It indicates that in real data there are some noise. Strong source of the noise is the intents which hard to distinguish. E.g. the phrase \cite{"I wand to go to Dublin"} may be for \emph{Find train}, \emph{Buy flight ticket}, \emph{Create route on the map} etc. To discard the influence of such confusions, we consider the metric $recall@4$.

%The best model for human Fluency score is random search with truncated on top-$0.6$ probability distribution with post-decoding clustering; it is quite close to human level on other metrics, but a bit less diverse. 

\begin{table*}[!htp]
\centering

\begin{tabular}
{p{8cm}p{5cm}p{0.75cm}p{0.75cm}}
%{|p{0.5\linewidth}|p{0.5\linewidth}|}
%\hline
\toprule
Intent description and reference examples & Undesirable meaning & Zero-shot & One-shot \\\midrule
{\bf Intent description} Train Buy train ticket \newline {\bf Reference} Make a purchase of the train ticket, not bus. Buy a train ticket for a specific date to some location &
{\bf Meaning} Get bus ticket \newline 
{\bf Example} I need a bus to go there. I need to leave on the 3rd of this month. & 97 & 23 \\\midrule
{\bf Intent description} Wallpapers Put default wallpaper \newline {\bf Reference} Change the background picture of the device display to the default one. Replace current background on the device with the default one &
{\bf Meaning} Put new wall cover in a house \newline
{\bf Example} I want to put the wallpaper for my bedroom on the wall. & 74 & 1 \\\midrule
{\bf Intent description} Calculator Find sum \newline {\bf Reference} Compute, calculate the sum of the given numbers. Open the calculator and compute the sum of the following numbers &
{\bf Meaning} Find some amount of money \newline {\bf Example} I need to find the average price of a house. & 57 & 0 
\\\bottomrule
\end{tabular}
\centering
\caption{Evaluation of semantic shift reduction by one-shot generation. The first column contains intent description and reference utterances used for one-shot generation. The second column shows examples of typical undesirable meaning. The last two columns show the percentage of examples with given incorrect meaning among 100 generated utterances by zero-shot and one-shot generation. Nucleus sampling ($p=0.4$) is used for both methods.}
\label{tab:ZHvsOS}
\end{table*}

\subsection{Semantic shift problem}
\label{sec:SemanticShift}
The semantic consistency is crucial: how well do the generated utterances correspond to the intent description? In most cases, zero-shot generation is quite reliable: $acc_{crowd}>0.8$ for $57\%$ of intents, $recall@4 > 0.9$ for $72\%$ of intents. However, generated utterances are distinguishable from other classes for some intents, but they do not completely correspond to the intent description. Several generated utterances below illustrate this issue.

{\bf Intent:} {\it Buy train tickets} \\
\noindent {\bf Utterance:} I want to buy a bus ticket. I want to leave on the 12th of this month. \\
{\bf Intent:} {\it Put default wallpapers} \\
\noindent {\bf Utterance:} Put the default wallpaper for the bedroom. I want to see it on the wall.\\
{\bf Intent:} {\it Calculator Find sum }\\
\noindent {\bf Utterance:} I need to find a calculator. I need to know the value of one dollar.

%All these examples are generated by the methods with relatively high accuracy ($ acc_{clsf} >80\%$). 
For example, The bias in the fine-tuning data causes this issue. For example, travel-related intents mainly correspond to bus travel. So the model confuses buses and trains. In other cases, the model gets wrong the intent description due to the lack of world knowledge. E. g. the generated phrases for {\it Wallpaper}  may be related to wallpapers in a house; utterances for {\it Calculator} may be related to finding some numbers like the average price of houses in the area. 

\iffalse
\begin{table}[h]
\caption{Semantic inconsistency examples.}
\centering
\begin{tabular}{ |l|l| }
\hline
\multicolumn{1}{|c|}{intent} & \multicolumn{1}{|c|}{phrase} \\
\hline
Wallpaper : Put default wallpaper &
Put the default wallpaper for the bedroom. I want to see it on the wall. \\
Trains : Buy train tickets &
I want to buy a bus ticket. I want to leave on the 12th of this month. \\
Calculator : Find sum &
I need to find a calculator. I need to know the value of one dollar. \\
Calculator : Count &
I want to know how many calories are in a cup of coffee \\
Call : Show phone number &
can you call my mom and ask for her number \\
\hline
\end{tabular}
\label{tab:semanticExamples}
\end{table}
\fi

\section{One-shot generation experiments}
Based on human evaluation of zero-shot generated data, we select Nucleus Sampling ($p=0.4$) as the best decoding strategy and apply it further in the one-shot scenario. Indeed, Table~\ref{tab:genComparison} confirms that the one-shot generation improves all evaluation metrics, both human and automated. The resulting one-shot utterances are more fluent than zero-shot utterances. The classifier trained on one-shot utterances has higher accuracy values when compared to the one trained on zero-shot utterances. 

At the same time, one-shot generation restricts the semantics of the generated utterances and reduces the semantic shift. To illustrate, how the problem of semantic shift diminishes, we study several cases where the zero-shot model tends to generate utterances with undesirable meaning (see Section~\ref{sec:SemanticShift}): {\bf bus} instead of {\bf train};  {\bf wallpaper} as a {\it wall cover} instead of {\it background picture}; {\bf sum} as {\it amount of money} instead of {\it number}. 
%For each of them, we generate utterances using nucleus sampling with $p=0.4$ in zero-shot and one-shot setup, 100 utterances per intent. 
Table~\ref{tab:ZHvsOS} shows that after one-shot fine-tuning, the number of utterances with undesirable meaning becomes drastically lower; for more examples, see Table 3 in Appendix.

\section{Conclusion}

In this paper, we have introduced zero-shot and one-shot methods for generating utterances from intent descriptions. We ensure the high quality of the generated dataset by a range of different measures for diversity, fluency, and semantic correctness, including a crowd-sourcing study. We show that the one-shot generation outperforms the zero-shot one based on all metrics considered. Using only a single utterance for an unseen intent to fine-tune the model increases diversity and fluency. Moreover, fine-tuning on a single utterance diminishes the semantic shift problem and helps the model gain better world knowledge. 

Virtual assistants in real-life setup should be highly adaptive. In some tasks, we need much more data than is currently available: exploring model robustness to distribution change, finding the best architecture, dealing with a fast-growing set of intents (the number of intents could be thousands). If the intents to support come from different providers, they pose diverse semantics, style, and noises. Adaptation to different user groups and individual users, having different intent usage distribution, is another crucial problem. We need large-scale and flexible datasets to approach these tasks, which can hardly be collected via crowd-sourcing from external sources.

Zero- or one-shot generation is an appealing technique. The model obtains the background knowledge about the world and the domain during pre-training. Next, only small amounts of data are needed to fine-tune the model. State-of-the-art pre-trained language models,  fine-tuned in a zero- or one-shot fashion, generate fluent and diverse phrases close to real-life utterances. The meaning of the intent and essential details, such as book titles, movie genres, expression of speech acts, or emoticons, are preserved.   What is more, manipulating a decoding strategy makes it possible to balance the generated utterances' diversity, semantic consistency, and correctness. 

Our future work directions include assessing the downstream performance of proposed generation methods for an end-user application and evaluating slot-filling performance. The proposed approach can be tested to generate utterances specific to interest groups.  

\section*{Acknowledgements}
Ekaterina Artemova is partially supported by the framework of the HSE University Basic Research Program.

%\newpage
\bibliography{anthology,emnlp2020}
\bibliographystyle{acl_natbib}

\end{document}

\clearpage
\onecolumn

\begin{table*}
\begin{tabular}{p{0.7\linewidth}|p{0.3\linewidth}}
\toprule
Generated utterance & Detected problem \\ \midrule
send to family members \emph{eldyiss} and me about \emph{okonomiyaki} & non-existing entities \\ 
please shut down the device. Detected missing Button on the self \emph{professile} webcam & non-existing words \\ 
in my contacts list name is cultural sound i am in with thirteen thousand six hundred and eighty seven & meaningless phrase \\
can you \emph{call my mom} and ask her to \emph{call my mom} & repetitions \\
I would like to find a bus for \emph{one people}. & grammar error \\
I'd like to find a \emph{bus flight}. & semantic error \\
I need help finding the \emph{average sum of two people}. & semantic error \\ 
\bottomrule

\end{tabular}
\caption{Fluency problems}
\label{tab:badFluencyExamples}
\end{table*}

\hspace{10em}

\begin{table*}[!hp]
\begin{tabular}{lccccccc}
\toprule
\multicolumn{1}{l}{Decoding strategy} & $acc_{clsf}$ & $Dist\mbox{-}1$ & $Dist\mbox{-}2$ & $Dist\mbox{-}4$ & $Ent\mbox{-}1$ & $Ent\mbox{-}2$ & $Ent\mbox{-}4$ 
\\\midrule
Beam Search ($b=1$) & 0.84  & 0.04  & 0.08  & 0.13  & 3.21  & 3.80  & 4.30 
\\
Beam Search ($b=1$)+PDC  & 0.84  & 0.06  & 0.13  & 0.21  & 3.42  & 4.14  & 4.75 
\\
Beam Search ($b=3$) & 0.87  & 0.04  & 0.08  & 0.13  & 3.11  & 3.76  & 4.35 
\\
Beam Search ($b=3$+PDC & 0.85  & 0.05  & 0.13  & 0.22  & 3.36  & 4.18  & 4.92 \\ 
Nucleus Sampling ($p=0.3$) & \bf{0.91}  & 0.05  & 0.10  & 0.14  & 3.15  & 3.72  & 4.16 \\
Nucleus Sampling ($p=0.4$)	& 0.89 	& 0.08 & 0.16 &	0.25 & 	3.45 &	4.26 &	4.95 \\
Nucleus Sampling ($p=0.6$) & 0.81  & 0.12  & 0.28  & 0.45  & 3.90  & 5.03  & 5.92 \\
Nucleus Sampling ($p=0.6$)+PDC & 0.82  & 0.10  & 0.24  & 0.40  & 3.86  & 4.90  & 5.77 \\
Nucleus Sampling ($p=0.8$) & 0.72  & 0.18  & 0.42  & 0.61  & 4.34  & 5.70  & 6.50 \\
Nucleus Sampling ($p=0.8$) +PDC & 0.75  & 0.14  & 0.34  & 0.52  & 4.15  & 5.36  & 6.20 \\
Nucleus Sampling ($p=0.98$) & 0.59  & \bf{0.26}  & \bf{0.59}  & \bf{0.72}  & \bf{4.84}  & \bf{6.32}  & \bf{6.86} \\
Nucleus Sampling ($p=0.98$)+PDC & 0.64  & 0.20  & 0.48  & 0.65  & 4.55  & 5.94  & 6.65 \\
Random Sampling ($k=3$) & 0.82  & 0.09  & 0.23  & 0.42  & 3.74  & 4.90  & 5.93 \\
Random Sampling ($k=3$)+PDC & 0.83  & 0.08  & 0.20  & 0.37  & 3.72  & 4.79  & 5.73 \\
Random Sampling ($k=4$) & 0.82 &	0.10 &	0.28 &	0.50 &	3.89 &	5.16 &	6.20 \\
Random Sampling ($k=8$) & 0.74  & 0.14  & 0.38  & 0.62  & 4.20  & 5.64  & 6.59 \\\bottomrule

\end{tabular}
\caption{Comparison of decoding strategies with different parameters using automated metrics of diversity and accuracy. Different diversity metrics correlate well. The highest diversity corresponds to the lowest accuracy.}
\label{tab:bigMetrics}
\end{table*}

\begin{table*}
\begin{tabular}{p{0.5\textwidth}|p{0.5\textwidth}}
\toprule
\multicolumn{1}{c|}{\bf{zero-shot}} & \multicolumn{1}{c}{\bf{one-shot}} \\ \midrule
\multicolumn{2}{l}{Intent description: \it{Device Turn on off bluetooth}} \\
\multicolumn{2}{l}{Reference: \it{Turn off bluetooth on phone device}}\\
\midrule
turn off my smart phone's wifi connection please.\newline
please turn off the on button for the speakers in the bedroom \newline
turn off my phone and turn off my tv remote \newline
turn off my smart phone for me please and please turn off the bluetooth device \newline
turn off my smart plug socket in my bedroom please &
Can you turn off the bluetooth device? I want to turn it on only when I'm in my bedroom. \newline
I would like to turn off the Bluetooth connection of my device. \newline
please turn off the bluetooth device in my home \newline
Yes, please turn off. Please make sure it is turned on in the bedroom. \newline
I would like to turn off Bluetooth. Please can you find the settings for this device? \\
\midrule
\multicolumn{2}{l}{Intent description: \it{Radio Turn on news}} \\
\multicolumn{2}{l}{Reference: \it{Turn on the news on the radio I want to know what is happening}}\\
\midrule
Please turn on the radio. Please play the latest news.\newline
I want to listen to the news on the radio.\newline
please turn on the radio for me to listen to the news\newline
Yes, please turn on the radio. I want to listen to the latest news.\newline
Yes, please turn on the news channel. I want to listen to the radio. &
That sounds good. Can you turn on the news? I want to hear about some interesting things happening in Washington D \newline
can you play the news radio for me please. I want to hear the president's press conference \newline
Yes, please turn on the news. I want to hear about the weather \newline
Please turn on the news. I want to hear about what's going around in politics \newline
Yes, please turn on the news. Please play it in the background. \\
\midrule
\multicolumn{2}{l}{Intent description: \it{Wallpapers Put default wallpaper}} \\
\multicolumn{2}{l}{Reference: \it{Change the background picture of the device display to the default one}}\\
\midrule
Put wallpaper for my bedroom. Put the wallpaper from the previous wallpaper \newline
I want to put my default wallpaper on my home screen \newline
I want to have the wallpaper set to my favorite wallpaper. \newline
Put wallpaper for me. I like Pop. Put it on my bedroom. \newline
I want to put the wallpaper for my bedroom on the wall. &
Sounds good. Please set the default wallpaper to the current theme. \newline
Thanks. Now please change the default wallpaper to "Quadriplegic". \newline
Okay, thanks. Can you put the default wallpaper? It's on the second. \newline
Yes, put the default wallpaper. Thanks for your help so far \newline
Sounds good. Please set the default wallpaper to Fullscreen. Default is shown in the settings. \\\hline
\multicolumn{2}{l}{Intent description: \it{Downloads on device Show downloads}} \\
\multicolumn{2}{l}{Reference: \it{Open the list of downloaded files on the current device}}\\\hline
I want to find a song. Can you find me some? \newline
show me the music from my music player list please \newline
i want to watch the movie on my device. \newline
show me the downloads for my phone and i have a device with android \newline
I want to find a song to listen to. &
That is great. Please show the downloaded files on my device as well \newline
% I would like to see the downloaded files on my device. I'd prefer if you show them in a list for me, please \newline
% Show downloads on device. i want to see the downloaded files for me. \newline
% Show downloads on device. It's not necessary to set up a regular display for the download. \newline
Okay, that's fine. Show me the downloaded files on my device \\ \bottomrule
\end{tabular}
\caption{Generated utterances for selected intents for both zero-shot and one-shot scenarios. The decoding strategy used is Nucleus Sampling (0.4) in both cases.  Top-$5$ generated samples for each model-intent  pair}
\label{tab:app_examplesZHvsOS}
\end{table*}

\begin{table*}
\resizebox*{!}{\textheight}{
% \small
\begin{tabular}{ p{0.15\textwidth} | p{0.85\textwidth}}

\toprule
Service & Intents \\ \midrule
Alarm\_1 & add alarm, get alarms\\
Banks\_1 & check balance, transfer money\\
Banks\_2 & check balance, transfer money\\
Buses\_1 & find bus, buy bus ticket\\
Buses\_2 & find bus, buy bus ticket\\
Calendar\_1 & add event, get events, get available time\\
Events\_1 & find events, buy event tickets\\
Events\_2 & find events, buy event tickets, get event dates\\
Flights\_1 & reserve oneway flight, search oneway flight, reserve roundtrip flights, search roundtrip flights\\
Flights\_2 & search oneway flight, search roundtrip flights\\
Flights\_3 & search oneway flight, search roundtrip flights\\
Homes\_1 & schedule visit, find apartment\\
Hotels\_1 & search hotel, reserve hotel\\
Hotels\_2 & search house, book house\\
Hotels\_3 & search hotel, reserve hotel\\
Hotels\_4 & search hotel, reserve hotel\\
Media\_1 & find movies, play movie\\
Media\_2 & find movies, rent movie\\
Movies\_1 & get times for movie, find movies\\
Movies\_2 & find movies\\
Music\_1 & lookup song, play song\\
Music\_2 & play media, lookup music\\
RentalCars\_1 & reserve car, get cars available\\
RentalCars\_2 & reserve car, get cars available\\
Restaurants\_1 & find restaurants, reserve restaurant\\
Restaurants\_2 & find restaurants, reserve restaurant\\
RideSharing\_1 & get ride\\
RideSharing\_2 & get ride\\
Services\_1 & find provider, book appointment\\
Services\_2 & find provider, book appointment\\
Services\_3 & find provider, book appointment\\
Services\_4 & find provider, book appointment\\
Travel\_1 & find attractions\\
Weather\_1 & get weather\\
alarm & set, remove, make query\\
audio & volume other, volume up, volume mute, volume down\\
calendar & set, remove, make query\\
cooking & recipe, make query\\
datetime & convert, make query\\
email & send message, make query contact, add contact, make query\\
general & ask to explain, nagate, stop command, dontcare, confirm, praise, affirm, greet, repeat, quirky, joke\\
iot & make a drink, cleaning, turn off, hue light on, turn on, hue light up, hue light dim, hue light off, hue light change\\
lists & create or add, remove, make query\\
music & like, settings, dislike, make query\\
news & make query\\
play & podcasts, play game, play audibook, radio, music\\
qa & stock, ask currency, ask definition, math, ask fact\\
recommendation & check events, movie, location\\
social & post, make query\\
takeaway & make order, make query\\
transport & get car, know traffic, buy ticket, make query\\
weather & make query\\ 
\bottomrule
\end{tabular}
}
\caption{Seen services and intents used for zero-shot fine-tuning. Intent descriptions were built by joining service name and intent name}
\end{table*}

% \begingroup
% \small
\begin{table}

\begin{longtable}{p{0.1\textwidth}p{0.15\textwidth}p{0.65\textwidth}}

\toprule
\multicolumn{1}{c}{Service name} & \multicolumn{1}{c}{Intent name} & Reference examples for one-shot generation \\
\midrule
% \endfirsthead
% % \hline
% \multicolumn{1}{c}{service name} & \multicolumn{1}{c}{intent name} & \multicolumn{1}{c}{reference examples for one-shot generation} \\
% \midrule
% \endhead
\multirow{4}{0.1\textwidth}{Alarm} & Create alarm & Set an alarm for tomorrow morning\\
  & Show alarm details & Show what time the alarm is set\\
  & Delete alarm & Delete the alarm set for a specific time\\
  & Cancel alarm & Cancel the alarm set on some time\\
  \midrule
\multirow{2}{0.1\textwidth}{App shop} & Buy app & Buy the app from the application store from the list of the most useful\\
  & Open lists of apps & Open apps lists in the application store\\
  \midrule
\multirow{2}{0.1\textwidth}{Audio Recorder} & Turn on record & Turn on the recorder and record everything that will be said\\
  & Stop record & Turn off the recorder and stop recording\\
  \midrule
\multirow{3}{0.1\textwidth}{Audio books} & Resume & Start playing a book, I want to listen to it\newline Keep playing audio book from where I left off\\
  & Search book & Find me a popular book in audio format\newline Search me an exciting book to listen to\\
  & Delete book & Remove an audio book that I listened to from device\newline Erase audio book, I don't want to listen to it\\
  \midrule
\multirow{2}{0.1\textwidth}{Books} & Find book & Look for a book of genre that I need and that interests me\newline Search me a literary work, preferably a book to read\\
  & Delete book & Delete book that are not interesting to me\newline Remove book from device that I downloaded last week and already read\\
  \midrule
\multirow{2}{0.1\textwidth}{Books reading list} & Create list & Make a list of books of my favorite genre\newline Create a list of all the books I have not read\\
  & Remove books & Delete lists of books that I have not read since last year\newline Remove last book reading list I created\\
  \midrule
\multirow{2}{0.1\textwidth}{Brightness} & Add brightness & Add brightness on the device screen\\
  & Reduce brightness & Decrease the brightness of the device screen\\
  \midrule
\multirow{2}{0.1\textwidth}{Calculator} & Count & Count the expression with a calculator\newline Calculate entered numeric expression\\\
  & Find sum & Compute, calculate the sum of the given numbers\newline Open the calculator and compute the sum of the following numbers\\\hline
\multirow{3}{0.1\textwidth}{Call on device} & Call & Call someone on the phone\\
  & Start video call & Call by video somebody from my contact list\\
  & Show phone number & Show the phone number of a particular person\\
\multirow{6}{0.1\textwidth}{Netflix} & Search movie & Find me an interesting movie of the desired genre\\
  & Play movie & Turn me on this movie online\\
  & Create playlist & Create a playlist of my favorite and most watched movies\\
  & Stop & Stop playing a movie\\
  & Resume & Start playing a film, I want to watch to it\\
  & Buy movie & Buy a movie that interested me\\
  \midrule
% \multirow{3}{0.1\textwidth}{Notes} & Start note & Start a note about what is happening\\\
%   & Delete note & Remove note created at some point\\
%   & Send note & Send my notes to me by mail\\\hline
% \multirow{3}{0.1\textwidth}{Payment} & Make payment & Pay for an order made by me earlier\\\
%   & Cancel payment & Cancel the order made by me earlier\\\
%   & Send money & Transfer money to a particular person\\\hline
% \multirow{2}{0.1\textwidth}{Plan trip} & Find hotel & Find me a hotel somewhere to stay for a few days\newline Look for a hotel so I can live there for a number of days\\\
%   & Book hotel & Book a hotel room in a certain place for a certain number of people\newline Rserve a room in a hotel for a number of days\\\hline
% \multirow{2}{0.1\textwidth}{Plane} & Buy plane ticket & Make a purchase of the plane ticket\newline Buy a plane ticket to some location\\\
%   & Find plane & Find a particular plane, for some date from one place to another\newline Find plane \\\hline
% \multirow{4}{0.1\textwidth}{Radio} & Turn on radio & Turn on the radio I want to listen to music or news\\\
%   & Turn on radio station & Turn on the radio station with my favorite radio programs\\\
%   & Turn on news & Turn on the news on the radio I want to know what is happening\\\
%   & Play music & Play music of a certain genre on the radio\\\hline
% \multirow{2}{0.1\textwidth}{Restaurant reservation} & Find restaurant & Find a restaurant for several people at the particular time in the right area\\\
%   & Reserve restaurant & Book table a the restaurant for several people\\\hline
\multirow{1}{0.1\textwidth}{Screenshot} & Make screenshot & Take a picture of the current screen of the device, make a screenshot\newline Make a screenshot of the device display\\
\bottomrule
\caption{A sample from the intent set for zero-shot and one-shot generation with reference examples. Intent descriptions were built by joining service name and intent name}
\label{tab:generationDataset}
\end{longtable}
\end{table}
% \endgroup

%\eegin{table*}[h]
%\centering
%\caption{Template}
%\centering
%\begin{tabular}{|l|c|ccc|ccc|}

%\end{tabular}
%\label{tab:templateTable}
%\end{table*}